\pdfoutput=1

\documentclass[11pt]{article}

\usepackage[final]{coling}

\usepackage{times}
\usepackage{latexsym}

\usepackage[T1]{fontenc}

\usepackage[utf8]{inputenc}

\usepackage{microtype}

\usepackage{inconsolata}

\usepackage{graphicx}
\usepackage{tikz}
\usetikzlibrary{arrows.meta, positioning}
\usepackage{booktabs}
\usepackage[linesnumbered,ruled,vlined]{algorithm2e}
\usepackage{amsmath}
\usepackage{amssymb}
\usepackage{CJKutf8}
\usepackage{hyperref}
\usepackage{float}

\title{CODEOFCONDUCT at Multilingual Counterspeech Generation: A Context-Aware Model for Robust Counterspeech Generation in Low-Resource Languages}

\author{
  \textbf{Michael Bennie}\textsuperscript{1},
  \textbf{Bushi Xiao}\textsuperscript{1},
  \textbf{Chryseis Xinyi Liu}\textsuperscript{1},
  \textbf{Demi Zhang}\textsuperscript{1},\\
  \textbf{Jian Meng}\textsuperscript{1},
  \textbf{Alayo Tripp}\textsuperscript{1}
  \\
  \textsuperscript{1}University of Florida, United States  \\
  \small{
    \textbf{Correspondence:} \href{mailto:michaelbennie@ufl.edu}{michaelbennie@ufl.edu}, \href{mailto:xiaobushi@ufl.edu}{xiaobushi@ufl.edu}, \href{mailto:liu.x1@ufl.edu}{liu.x1@ufl.edu}, \href{mailto:zhang.yidan@ufl.edu}{zhang.yidan@ufl.edu}
  }
}

\begin{document}
\maketitle
\begin{abstract}
This paper introduces a context-aware model for robust counterspeech generation, which achieved significant success in the MCG-COLING-2025 shared task. Our approach particularly excelled in low-resource language settings. By leveraging a simulated annealing algorithm fine-tuned on multilingual datasets, the model generates factually accurate responses to hate speech.

We demonstrate state-of-the-art performance across four languages (Basque, English, Italian, and Spanish), with our system ranking first for Basque, second for Italian, and third for both English and Spanish. Notably, our model swept all three top positions for Basque, highlighting its effectiveness in low-resource scenarios.

Evaluation of the shared task employs both traditional metrics (BLEU, ROUGE, BERTScore, Novelty) and JudgeLM based on LLM. We present a detailed analysis of our results, including an empirical evaluation of the model performance and comprehensive score distributions across evaluation metrics.

This work contributes to the growing body of research on multilingual counterspeech generation, offering insights into developing robust models that can adapt to diverse linguistic and cultural contexts in the fight against online hate speech.
\end{abstract}

\section{Introduction}



Hate speech (HS) encompasses expressions that demean or target individuals or groups based on characteristics such as race, ethnicity, gender, sexual orientation, or religion \citep{de-gibert-etal-2018-hate}. Although HS makes up only a small fraction of social media content, its impact is significant, affecting nearly one-third of people \citep{vidgen-etal-2019-challenges}. The prevalence of HS on social media has become a critical societal concern. Traditional approaches, such as content removal and moderation, have been widely implemented but are often criticized for infringing on free speech. As an alternative, counterspeech (CS) has emerged as a promising solution to mitigate HS while upholding the principle of free expression \citep{poudhar-etal-2024-strategy}.  

Counterspeech is defined as speech intended to counteract and neutralize harmful language. It has demonstrated effectiveness in real-world applications \citep{https://doi.org/10.1111/phc3.12890}, but its manual creation is labor-intensive and impractical at scale given the high volume of HS online. This challenge has driven interest in automating CS generation using NLP technologies. The growing need for effective counterspeech highlights the importance of strategies that foster healthier online environments. However, multilingual CS generation remains a significant challenge, particularly in low-resource settings where data scarcity limits model development. Research has focused on understanding HS, crafting effective CS, and addressing the unique challenges of multilingual and resource-constrained contexts.

The MCG-COLING-2025 shared task addressed these challenges by inviting researchers to generate respectful, specific, and truthful counterspeech across multiple languages, including Basque, English, Italian, and Spanish. In this paper, we present *CODEOFCONDUCT*, a context-aware model that achieved state-of-the-art performance in this shared task. Our model demonstrates exceptional effectiveness in low-resource scenarios, particularly for Basque, and offers valuable insights into leveraging multilingual datasets and advanced optimization techniques for counterspeech generation.

\vspace{1em}
\noindent \textbf{Our contributions}
\begin{itemize}
    \item A novel application of the simulated annealing algorithm in fine-tuning for counterspeech generation.
    \item Competitive performance across four languages, including top-ranking results in low-resource language Basque.
    \item Critical analysis of provided CS evaluation tools, especially Judge-LM.
\end{itemize}



\section{Background}
The languages chosen for the shared task cover a wide range of linguistic features, from Basque’s complex agglutinative morphology to the more straightforward syntax of English. This diversity allows for testing models' adaptability to varying linguistic challenges. Furthermore, the inclusion of background knowledge and cultural differences in multiple languages adds an additional layer of complexity, requiring models to integrate contextual information effectively.
\subsection{Opportunities and Gaps}
Multilingual counterspeech generation is challenging due to variations in linguistic structure, cultural norms, and resource availability. English dominates this field due to its resource-rich environment, while languages like Basque lack sufficient annotated data and pretrained models (\citet{faisal2021lowresource}). Approaches to address these challenges often leverage transfer learning and multilingual pretraining. Models like mBERT (\citet{devlin2019bert}) and XLM-RoBERTa (\citet{conneau2020xlmroberta}) have demonstrated robust cross-lingual transfer capabilities, enabling better performance in low-resource settings. Fine-tuning these models on task-specific data, such as the HS-CN pairs in this shared task, has shown promise in generating meaningful and contextually relevant counterspeech.

Notably, the integration of background knowledge to enhance counterspeech quality has been explored in previous works (\citet{qian2019benchmark}), which introduced external knowledge to improve the informativeness of responses. This concept is particularly relevant in the MCG-COLING-2025 task that utilizes the CONAN dataset, where models must use hate speech, background knowledge, and linguistic nuances to synthesize counter speech.

\subsection{Low-Resource NLP: Basque as a Case Study}
To bridge these gaps, which are due to their complex morphology and limited linguistic resources, techniques such as data augmentation \citep{feng2021survey}, transfer learning, and multilingual pretraining have been explored. A Basque BERT model \citep{agerri-etal-2020-give} has been developed using a dataset drawn from the Basque edition of Wikipedia and news articles from various Basque media sources, illustrating the potential of specialized models in low-resource contexts.

For counterspeech generation, low-resource languages require innovative solutions to overcome data scarcity. Judge-EUS\footnote{\href{https://huggingface.co/HiTZ/judge-eus}{https://huggingface.co/HiTZ/judge-eus}} has been utilized to enhance response quality, as demonstrated below in our approach.

\subsection{Multilingual Hate Speech-Counter Narrative Dataset}
For this shared task, the training dataset involves 596 Hate Speech-Counter Narrative (HS-CN) pairs curated by the task organizers. The hate speech instances were sourced from the Multitarget-CONAN dataset \cite{fanton-etal-2021-human}, and the counterspeech responses were newly generated by the organizers. Each HS-CN pair is accompanied by five background knowledge sentences, some of which are relevant for crafting effective counterspeech.

The dataset spans four languages: Basque, English, Italian, and Spanish, representing a diverse typological spectrum:

\begin{itemize}
    \item \textbf{Basque}: An agglutinative language isolate with unique grammatical structure.
    \item \textbf{Italian and Spanish}: Two Romance languages with high lexical and syntactic similarity.
    \item \textbf{English}: A Germanic language with relatively simpler morphological structures.

\end{itemize}

The dataset is divided into the following splits for each language:
\begin{itemize}
    \item \textbf{Development}: 100 HS-CN pairs.
    \item \textbf{Training}: 396 HS-CN pairs.
    \item \textbf{Testing}: 100 HS-CN pairs (counter-narratives held out as blind test data).
\end{itemize}


\subsection{Current Evaluation Metrics}

\begin{itemize}
   
 \item JudgeLM: An LLM-based ranking method for evaluating automatic counter-narrative generation \citep{zubiaga2024llmbasedrankingmethodevaluation}.
 \item BLEU: Measures token overlap between predictions and references \citep{papineni2002bleu}.
 \item ROUGE-L: Computes sentence-level structure similarity and longest co-occurring n-grams \citep{lin2004rouge}.
 \item BERTScore: Calculates token-level similarity using contextual embeddings \citep{DBLP:journals/corr/abs-1904-09675}.
 \item Novelty: Measures the proportion of non-singleton n-grams in generated text that do not appear in the training data \citep{Wang2018SentiGANGS}.
 \item Genlen: The average length of generated predictions.\end{itemize}

\section{Methodology}

We employed a three-stage approach for generating effective CS across multiple languages. The first stage utilizes a simulated annealing approach combined with LLMs to generate and select diverse responses. A word sampling mechanism extracts vocabularies from both predefined word lists and input text (HS and CN) to enrich the response generation. Each candidate response is evaluated using JudgeLM, with scores exponentially weighted to guide the sampling process. The second stage implements a Round-robin tournament evaluation system to rank and select the most effective responses, ensuring high-quality output even in low-resource language settings. In the last stage, we combined each of the first-ranked, second-ranked, third-ranked, and fourth-ranked answers into their own CSVs and then ran the evaluation script 
given by MCG-COLING \footnote{\url{https://github.com/hitz-zentroa/eval-MCG-COLING-2025/blob/master/evaluation/bash/judge_full_pipeline.sh}} to find the top 3 runs.

\subsection{Comparison to other methods}
Our approach of using JudgeLM to rank candidate counter-narratives (CNs) parallels prior methods proposed by \citet{Zubiaga2024IxaAR}, where Large Language Models (LLMs) generate CNs and an LLM-based evaluator selects the best response via pairwise comparisons. While both systems rely on tournament-style evaluation using an LLM judge, our framework fundamentally diverges by incorporating a simulated annealing stage before the round-robin tournament to generate a set of iteratively refined answers to compare. Instead of generating each CN in a single pass, we repeatedly mutate, expand, and re-score candidate responses, enabling a broader exploration of the CN space and reducing the risk of local optima. By adjusting a temperature parameter, even lower-scoring CNs can remain viable candidates at early stages, fostering globally stronger outputs. Furthermore, we augment vocabulary sampling with tokens from both predefined lists and the original hate speech to try to create contextually grounded answers. 

\subsection{Data Used}

As our model did not require training or any other outside data, we elected to only use the testing set (100 HS-CN pairs for each language) of the data provided to generate our answers. It was determined that the use of the development and training subsets of data was not necessary as we could directly test the quality of generated answers using JudgeLM.

\subsection{Stage 1: Counterspeech Generation}

In this stage, we implemented a simulated annealing algorithm (Algorithm~\ref{alg:simulated-annealing} from the appendix) to generate effective counterspeech (CS) responses to hate speech (HS) instances across multiple languages. The algorithm iteratively refines CS candidates by exploring the search space in a manner inspired by thermodynamic annealing processes.

The algorithm begins with an initial CS candidate $c_0$, which can be the HS instance $h$ itself or another string. In our case, we used the background knowledge provided with each HS-CS pair from MCG-COLING as the initial string. At each iteration, we update the temperature parameter $T$ by an increment $\Delta T$, controlling the exploration-exploitation trade-off.

For each candidate CS $c$ in the current set $C$, we generate a set of new candidates $S$ by appending randomly sampled words from a language-specific word list. Notably, part of the vocabulary that we sample from is from the tokenized HS. This sampling enriches the vocabulary and introduces relevant words from the original HS in the candidate responses.

We evaluate each candidate $c' \in S$ using an LLM-based judge to obtain a score $E(c')$, reflecting its relevance, fluency, and effectiveness as a counterspeech. To prioritize higher-quality candidates while still allowing exploration of the search space, we compute selection probabilities using a Boltzmann-like distribution (Algorithm~\ref{alg:compute_probabilities} from the appendix):

\[
P(c') = \frac{T^{E(c')}}{\sum_{c'' \in S} T^{E(c'')}}.
\]

This probability distribution ensures that candidates with higher scores are more likely to be selected, but candidates with lower scores still have a chance of being chosen, especially at lower temperatures. This mechanism allows the algorithm to avoid local optima by occasionally exploring less promising candidates.

We select $k$ candidates from $S$ based on the computed probabilities $P(c')$. For each selected candidate $c'$, we generate new CS responses $\tilde{S}$ using Language Models (LLMs). 
These LLM-generated responses further diversify the candidate pool and introduce potentially high-quality CS that may not be reachable through simple word appending.
The exact LLMs used for counterspeech generation can be found in the appendix in Table \ref{tab:models}. 

The new candidates $\tilde{c} \in \tilde{S}$ are evaluated, and their probabilities are computed in the same manner. We update the candidate set $C$ with the top candidates from $\tilde{S}$ based on their probabilities. This process is repeated for a predefined number of iterations or until a candidate reaches the target score $S_{\text{target}}$. 

We optimized these hyper-parameters by experimenting on a small subset of 4 HS instances (one from each language) and measuring the average high score achieved. The results of this hyper-parameter tuning are presented in Table~\ref{tab:hyperparameter_tuning} in the Appendix. We observed that increasing the number of iterations and candidates per loop improved the average high score, with the combination of 8 iterations and 6 candidates per loop achieving an average high score of 10, which meets our target score threshold. 

\begin{figure}[h]
    \centering
    \includegraphics[width=1\linewidth]{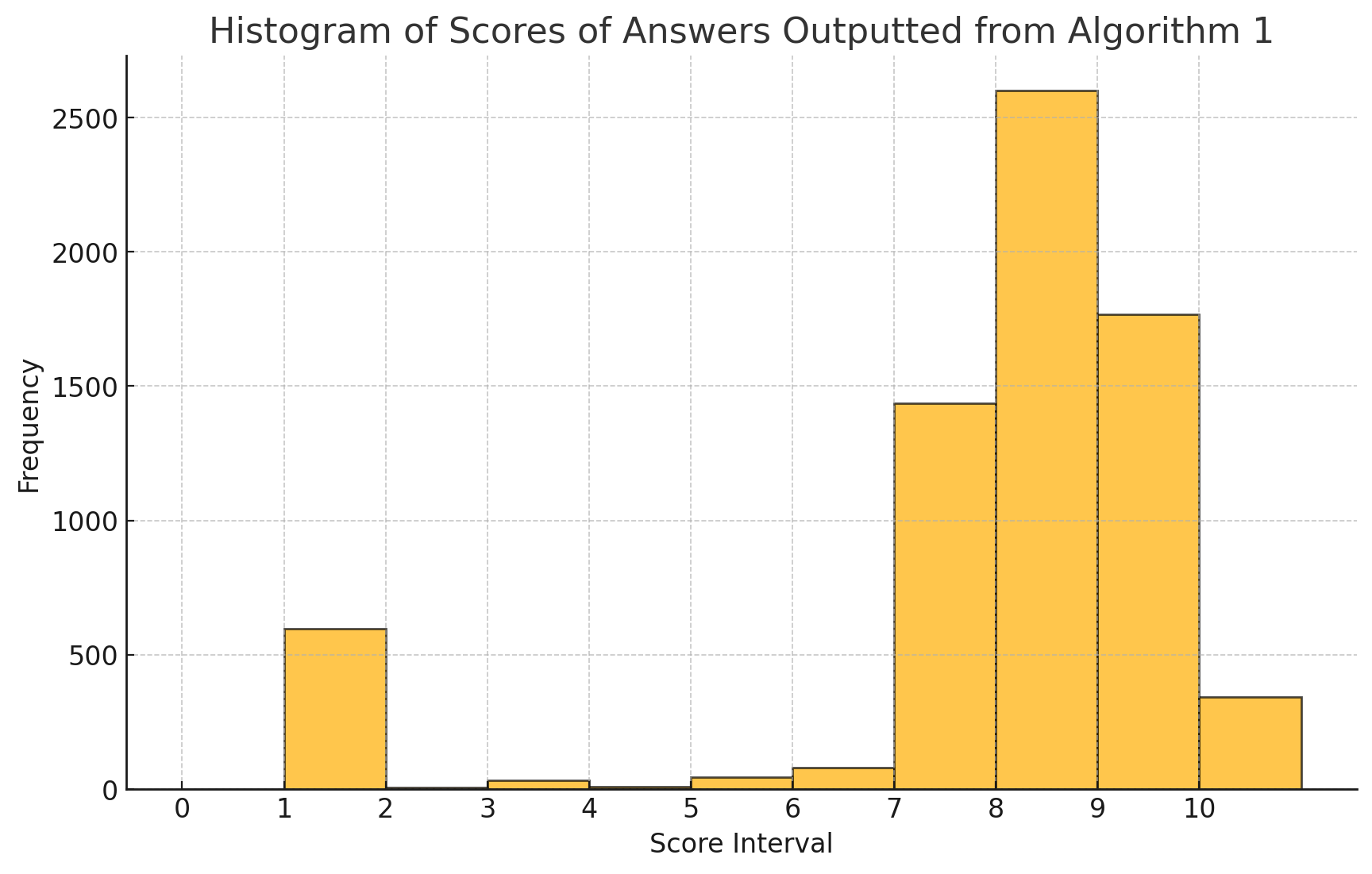}
    \caption{A histogram of scores $E(c)$ for every counterspeech generated from Algorithm \ref{alg:simulated-annealing}}.
    \label{fig:init-score-hist}
\end{figure}

We used these optimized parameters to generate sentences for MCG-COLING 2025 where we set the target scores of each CS answer ($S_{target}$) to be 9. After generating answers for each of the 400 total instances of hate speech, we were left with a set of a 6,915 answers. The distribution of the initial scores for each answer can be found in Figure \ref{fig:init-score-hist}. As can be seen, most answers achieved scores of 8 or higher from JudgeLM at this stage.

Assuming each iteration has $k$ CS options, each of which makes $k$ LLM generation calls that produce $n$ answers, the worst-case time complexity of this algorithm would be $O(N_{max}\cdot n\cdot k^2)$ calls to JudgeLM.

\subsection{Stage 2: Round Robin Ranking}
\begin{figure}
    \centering
    \includegraphics[width=1\linewidth]{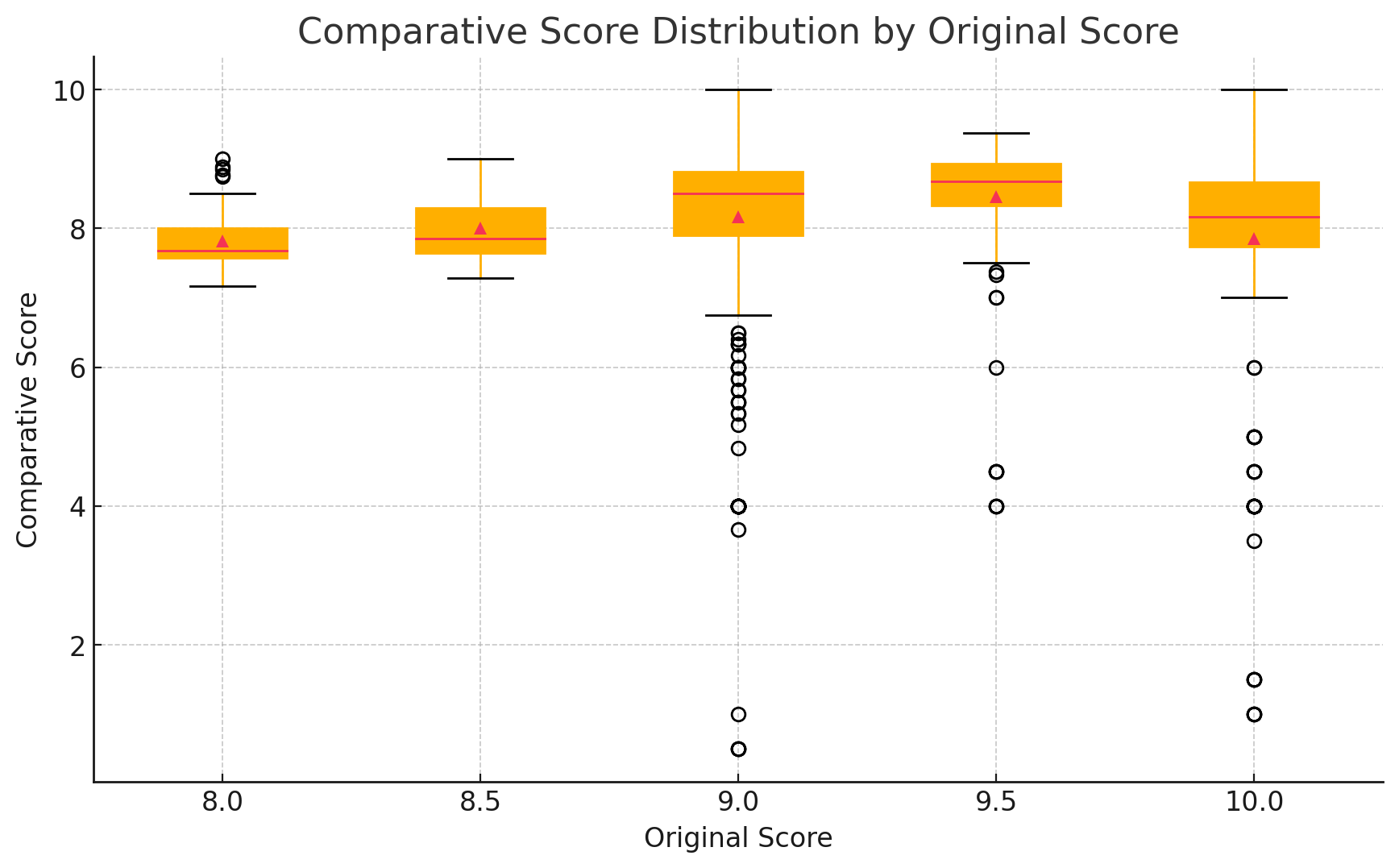}
    \caption{Box and whisker charts that compare the original scored value of a CS answer from stage 1 to the re-scored vales from stage 2.}
    \label{fig:RRanswerComparisons}
\end{figure}

While the simulated annealing algorithm can consistently generate answers that score a 10, these answers are not necessarily the best possible responses. Since JudgeLM comparatively ranks answers, the score of an answer can be affected by the quality of the other candidates' answers. This phenomenon is exemplified in Figure~\ref{fig:RRanswerComparisons}, which shows that there is a large variance (especially among high-scoring answers) between the score a specific counterspeech ($c$) received in the original algorithm ($E(c)$) and the score it received when recalculated in comparison to other high-scoring answers ($\text{RoundRobin}(c, C)$) where $C$ is a set of possible CS answers for a given HS.

Ideally, we would compare every answer with every other answer at the end of each loop of the simulated annealing algorithm to identify the best responses. However, repeatedly performing these comprehensive comparisons in each iteration would cause the time complexity to skyrocket to $O(N_{\text{max}} \cdot n^2 \cdot k^4)$, where $N_{\text{max}}$ is the maximum number of iterations, $n$ is the number of candidates per loop, and $k$ is the number of candidates selected for further exploration. Such computational demands are impractical for our application.

Therefore, we implement a post-hoc algorithm to compare all the generated high-scoring answers and determine which ones are the best. The round-robin algorithm (Algorithm~\ref{alg:round-robin} from the appendix) addresses this by taking the final set of high-scoring answers from Algorithm~\ref{alg:simulated-annealing} and assigning each an average score based on pairwise comparisons. Each response in the answer pool is evaluated against every other response using JudgeLM, creating a tournament-style evaluation where each pair of responses is compared twice by swapping their positions to reduce position bias. As this algorithm has a quadratic time complexity, we only used the top 6 answers in the calculation. These comparison scores are accumulated and averaged across all matches and then divided by the total number of matches to generate an average scoring. This was used to provide a robust overall ranking for each counterspeech response. The top-four ranked answers were then stored.

\subsection{Stage 3: Generation of Submission Files}

Upon completion of the ranking process, four files (one for each rank of answer) were generated for each language's answers. A final sanity check was completed for each language by comparing each of the 4 files with a round-robin based scoring algorithm from MCG-COLING. For English, Spanish, and Italian, results were consistent with expectations; files containing higher ranking answers received a higher score. As seen in Figure \ref{fig:BasuqeScore}, run 4, despite being made of 4th place answers, scored 3.5 points higher than the file for the second rank answers. As such, the final submission for Basque in the 2025 MCG-COLING task contained the first, second, and fourth rank answers.

\section{Conclusion}

\begin{figure*}
    \centering
    \includegraphics[width=1.07\linewidth]{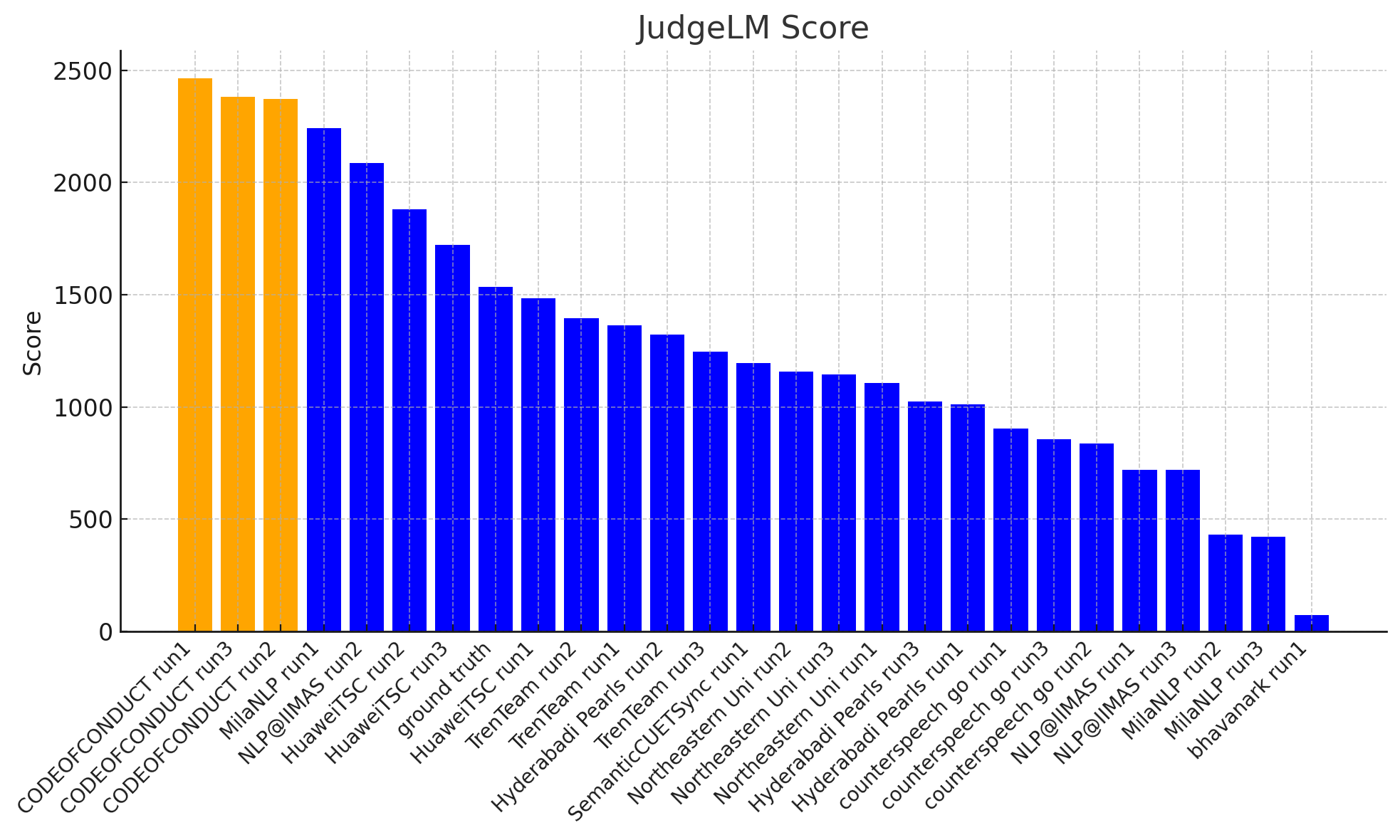}
    \caption{Chart depicting the JudgeLM scores for each Basque run. Bars drawn in yellow represent the results from the CODEOFCONDUCT submission.}
    \label{fig:basque-scores}
\end{figure*}

This study showcases the effectiveness of the CODEOFCONDUCT model in crafting context-aware counterspeech, achieving exceptional results in both high- and low-resource language settings. By integrating a simulated annealing-based generation framework with a robust round-robin ranking mechanism, our approach secured leading positions in the MCG-COLING-2025 shared task across four diverse languages.

As evidenced by the quantitative results in Figure \ref{fig:BasuqeScore}, our model’s success with Basque—a language with limited NLP resources such as annotated datasets, pretrained models, and linguistic tools—stands out as a key achievement. This success stems from three key factors: (1) our simulated annealing approach's effectiveness in handling Basque's complex agglutinative morphology, (2) a word sampling strategy specifically enhanced for low-resource scenarios by incorporating domain-specific terminology, and (3) an evaluation system well-suited to Basque's unique linguistic features. All three of the runs submitted were able to outclass all the other runs submitted by other groups. By working with only 100 HS-CN pairs for each language and leveraging multilingual pretrained models alongside innovative optimization techniques, we demonstrated how thoughtful methodologies can overcome resource constraints. Our sweeping success in Basque, coupled with strong rankings in Italian, English, and Spanish, highlights the versatility of our approach in navigating diverse linguistic and cultural challenges, with particular effectiveness in tackling the unique demands of low-resource languages.

Through the use of comprehensive evaluation metrics, including JudgeLM and traditional measures such as BLEU and ROUGE, we ensured that the generated counterspeech was both linguistically accurate and contextually appropriate. The combination of simulated annealing and round-robin evaluation is particularly well-suited for this task for several reasons: simulated annealing enables exploration of culturally appropriate responses through temperature-controlled sampling, while round-robin evaluation captures the nuanced effectiveness of counterspeech across different cultural contexts that simple metrics might miss. The final results highlighted the model's strengths while also revealing opportunities for improvement, particularly in adapting evaluation frameworks to better reflect the nuances of multilingual and culturally specific outputs.

\section{Limitations}

\subsection*{Static Response Generation}

Our CS generation system, while effective in current contexts, faces inherent temporal limitations. The greedy annealing algorithm's reliance on static word lists and predefined evaluation metrics constrains its ability to adapt to rapidly evolving hate speech patterns. As \citet{lupu2023offline} highlighted, hate speech can shift dramatically with offline events, challenging our model's static approach. \citet{DBLP:journals/corr/abs-2004-01670} emphasized how the "garbage in, garbage out" principle affects such systems, highlighting the need for more dynamic algorithmic solutions that can adapt to emerging patterns in real-time.

\subsection*{Local Optimality}

While the simulated annealing approach ensures rapid initial solution generation and efficient task distribution, it may not achieve global optimality. This occurs when the temperature decreases too rapidly or when the initial sampling conditions restrict exploration of the full solution space. Although we mitigate these issues through parameter tuning, the fundamental trade-off between exploration and exploitation remains a challenge.

\subsection*{Computational Cost}
The method of combining simulated annealing and round-robin evaluation introduces significant computational cost. Using an NVIDIA A100 GPU, processing one language's test set requires approximately 10 hours of computation time, primarily due to two factors: (1) the quadratic number of calculations needed for each simulated annealing iteration, and (2) the quadratic time complexity of round-robin evaluation required for generating high-quality responses. High computational cost makes it challenging to meet the demands of real-world applications.

\subsection*{Evaluation Metrics}
While JudgeLM demonstrates competence in English response evaluation, its lack of fine-tuning on multilingual counterspeech data affects its reliability. Basque responses require a separate Judge-EUS model, leading to potential inconsistencies in evaluation standards across languages. These models may overemphasize lexical similarities while missing cultural nuances and language-specific expressions, potentially leading to responses that score well numerically but fail to resonate with speakers of non-English languages.

\bibliography{custom}

\appendix
\label{sec:appendix}
\newpage
\section{Appendix}
\subsection{Algorithms}

\begin{algorithm}[]
\SetAlgoLined
\KwData{Hate speech $h$, initial counterspeech $c_0$, target score $S_{\text{target}}$, max iterations $N_{\max}$,candidates per loop $k$, temp increment $\Delta T$, initial temp $T_0$}
\KwResult{Optimal counterspeech $c^*$}

Initialize $C \leftarrow \{ c_0 \}$, $T \leftarrow T_0$\;
\For{$i = 1$ \KwTo $N_{\max}$}{
    $T \leftarrow T + \Delta T$\;
    $C_{\text{new}} \leftarrow \emptyset$\;
    \ForEach{$c \in C$}{
        Generate candidates $S$ by appending random words to $c$\;
        Evaluate scores $E(S)$ using LLM judge\;
        Compute probabilities $P(S)$ using Algorithm~\ref{alg:compute_probabilities}\;
        Select $k$ candidates from $S$ based on $P(S)$\;
        Generate new counterspeeches $\tilde{S}$ using LLMs on selected candidates\;
        Evaluate scores $E(\tilde{S})$\;
        Compute probabilities $P(\tilde{S})$\;
        Add top $k$ candidates from $\tilde{S}$ to $C_{\text{new}}$ based on $P(\tilde{S})$\;
    }
    Update $C \leftarrow C_{\text{new}}$\;
    \If{$\exists\, c \in C$ such that $E(c) \geq S_{\text{target}}$}{
        \Return $c^*$ with highest $E(c)$ in $C$\;
    }
}
\Return $c^*$ with highest $E(c)$ in $C$\;

\caption{Simulated Annealing for Counterspeech Generation}\label{alg:simulated-annealing}
\end{algorithm}

\begin{algorithm}[]
\SetAlgoLined
\KwData{Candidates $S$, scores $E(S)$, temperature $T$}
\KwResult{Probabilities $P(S)$}
\ForEach{$c' \in S$}{
    Compute probability:
    \[
    P(c') = \frac{T^{E(c')}}{\sum_{c'' \in S} T^{E(c'')}}
    \]
}
\caption{Compute Probabilities}\label{alg:compute_probabilities}
\end{algorithm}

\begin{algorithm}[]
\SetAlgoLined
\KwData{Counterspeech $c$, set of other counterspeeches $C$}
\KwResult{Average score of $c$}

Initialize $total\_score \leftarrow 0$\;

\ForEach{$a \in C$}{
    Create question comparing $c$ and $a$\;
    
    $normal\_results \leftarrow$ Evaluate $c$ vs.\ $a$ (normal order) using JudgeLM\;
    
    $reversed\_results \leftarrow$ Evaluate $a$ vs.\ $c$ (reversed order) using JudgeLM\;
    
    $score_c \leftarrow normal\_results["output1"] + reversed\_results["output2"]$\;
    $total\_score \leftarrow total\_score + score_c$\;
}
Compute $average\_score \leftarrow total\_score / (2 \times |C|)$\;

\Return $average\_score$\;

\caption{Round Robin Ranking for Counterspeech Evaluation}
\label{alg:round-robin}
\end{algorithm}

\subsection{Data}

\begin{figure}[H]
    \centering
    \includegraphics[width=1\linewidth]{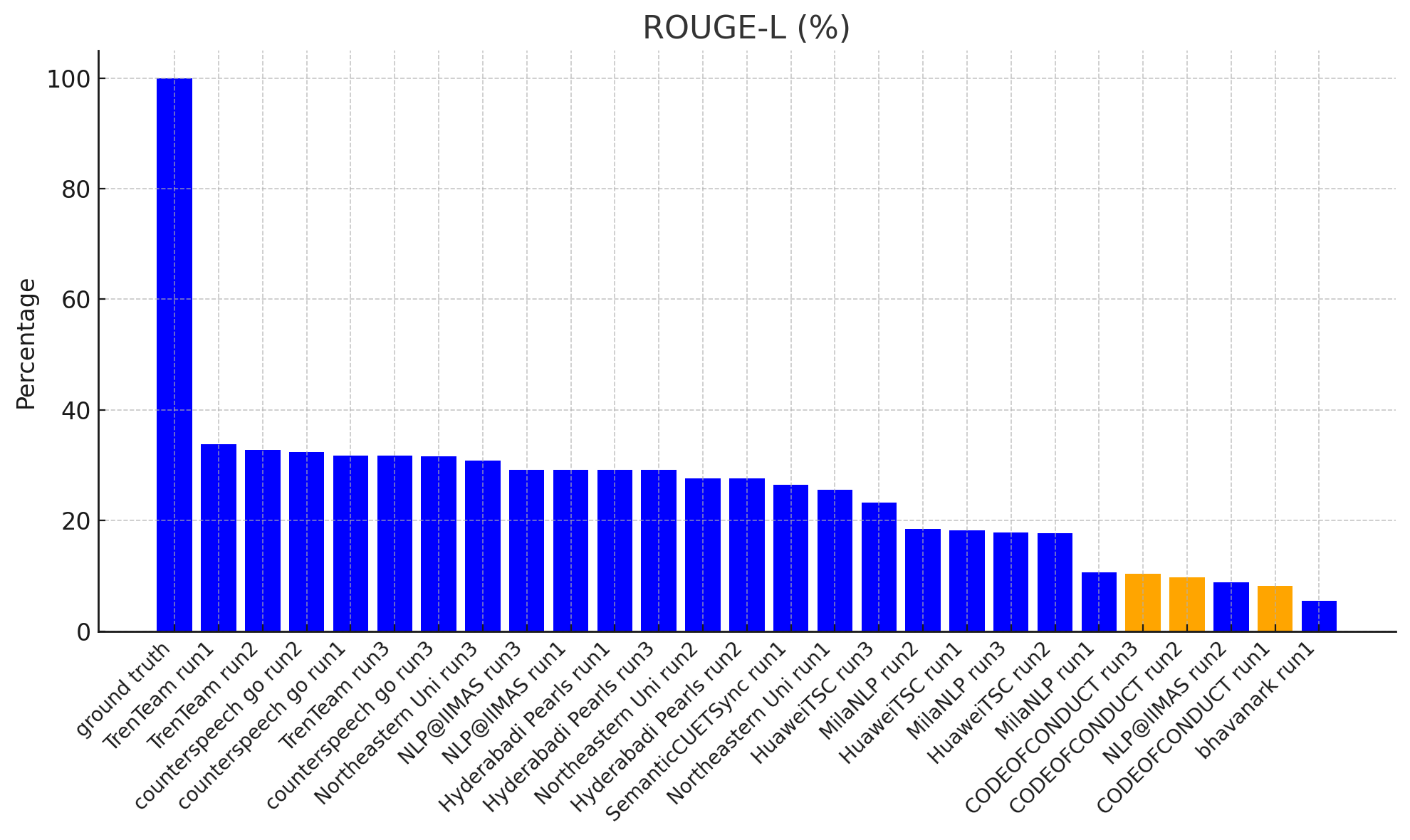}
    \caption{Chart depicting the ROUGE-L scores for each Basque run. Bars drawn in yellow represent the results from the CODEOFCONDUCT submission.}
    \label{fig:rougel}
\end{figure}

\begin{figure}[H]
    \centering
    \includegraphics[width=1\linewidth]{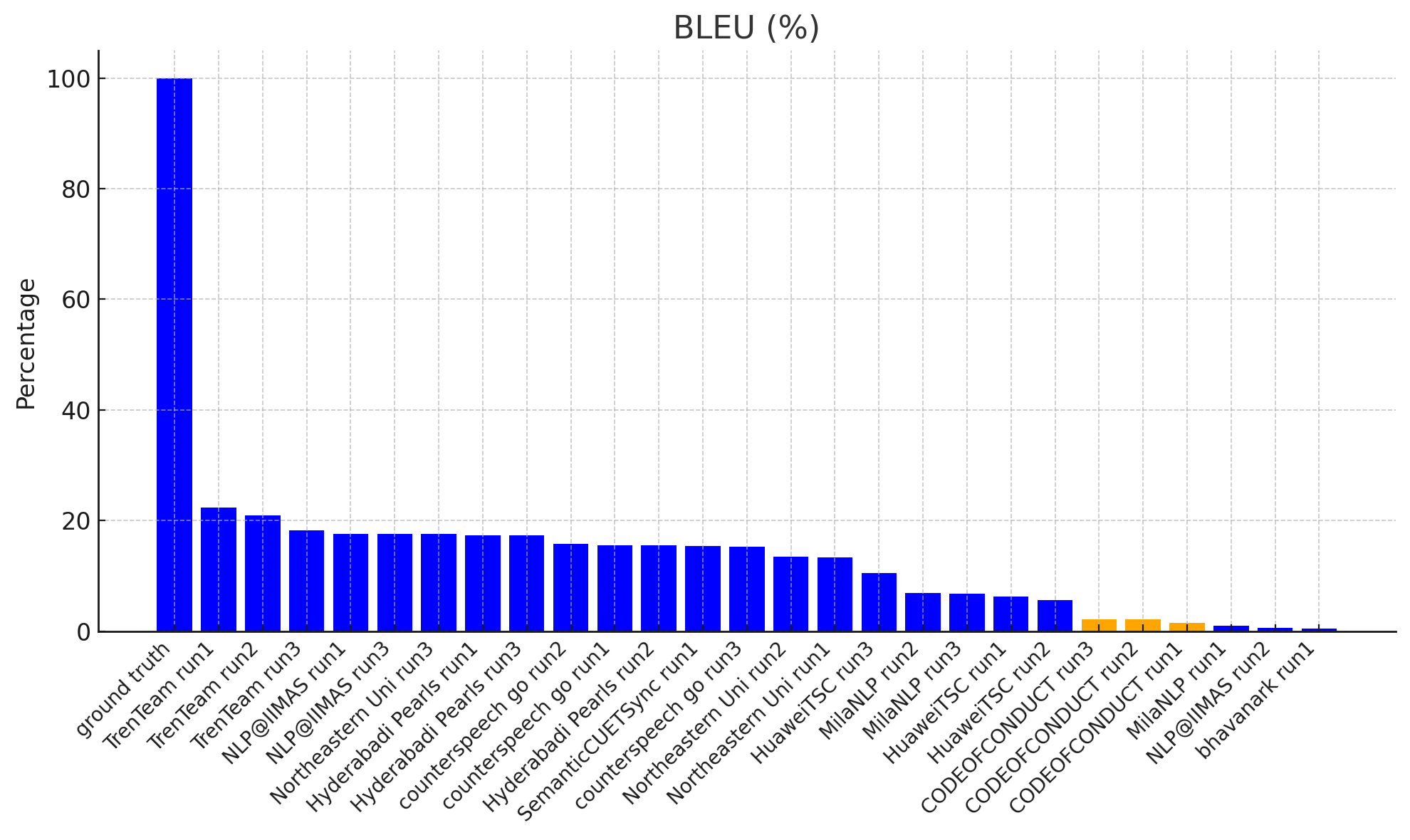}
    \caption{Chart depicting the BLEU scores for each Basque run. Bars drawn in yellow represent the results from the CODEOFCONDUCT submission.}
    \label{fig:bleu}
\end{figure}

\begin{figure}[H]
    \centering
    \includegraphics[width=1\linewidth]{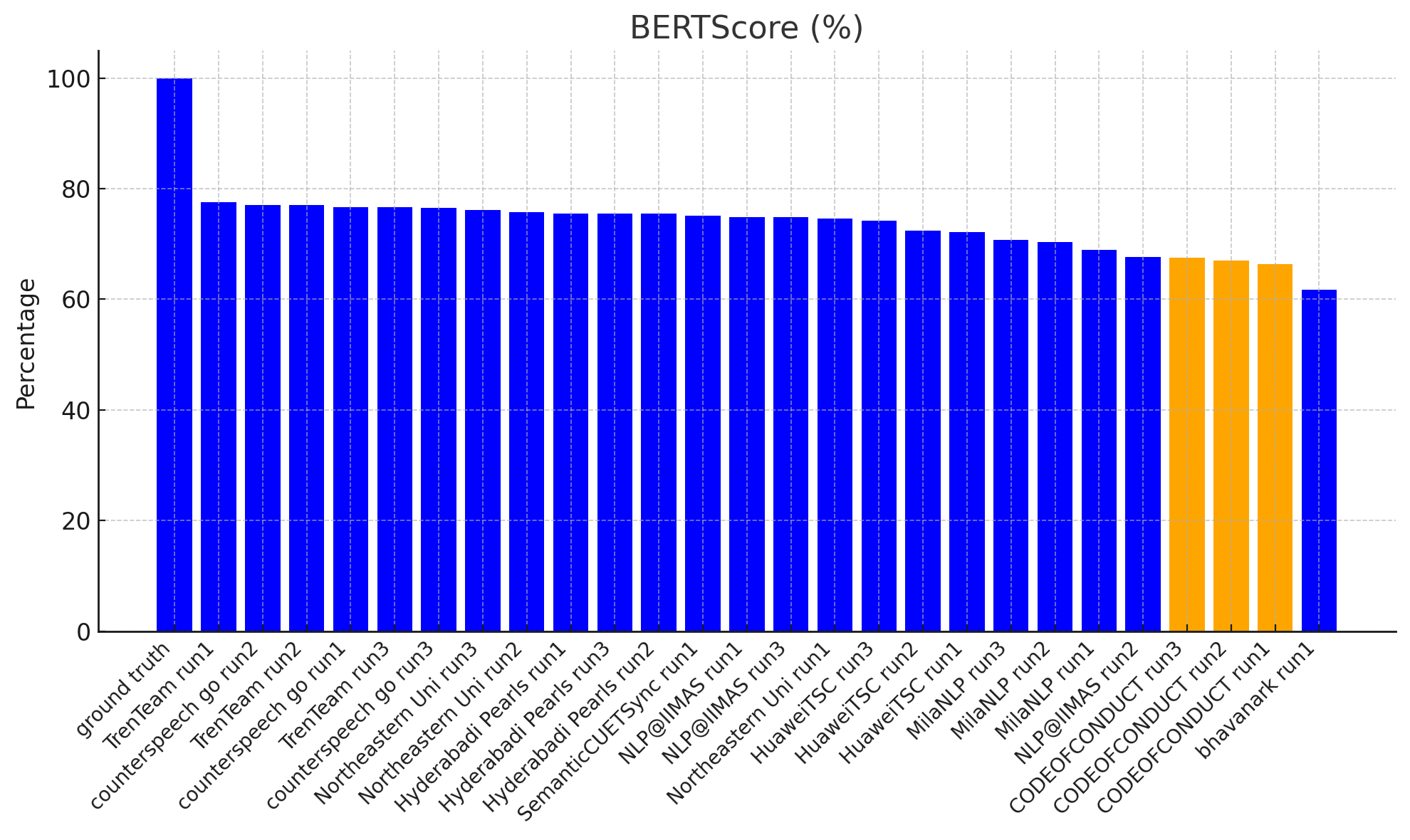}
    \caption{Chart depicting the BERT scores for each Basque run. Bars drawn in yellow represent the results from the CODEOFCONDUCT submission.}
    \label{fig:Bert}
\end{figure}

\begin{figure}[h]
    \centering
    \includegraphics[width=1.05\linewidth]{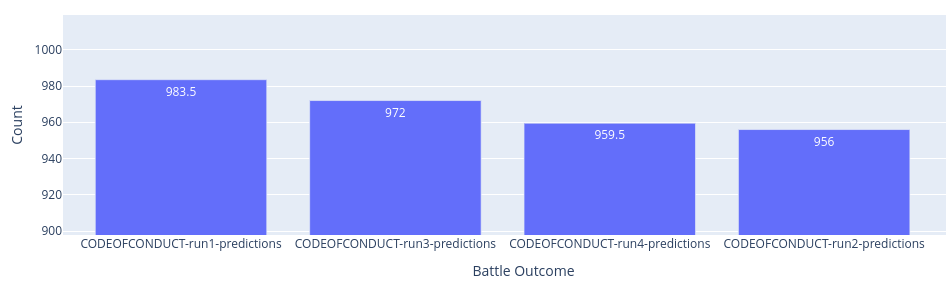}
    \caption{Scoring that compares the 4 runs of generated HS-CS files for Basque}
    \label{fig:BasuqeScore}
\end{figure}

\begin{table}[H]
\centering
\begin{tabular}{ccc}
\toprule
\textbf{Iterations} & \textbf{Candidates per Loop} & \textbf{Average High Score} \\
\midrule
2 & 2 & 8.875 \\
4 & 2 & 9.375 \\
6 & 2 & 9.000 \\
8 & 2 & 9.500 \\
\midrule
2 & 4 & 9.125 \\
4 & 4 & 9.875 \\
6 & 4 & 9.875 \\
8 & 4 & 9.500 \\
\midrule
2 & 6 & 9.625 \\
4 & 6 & 9.625 \\
6 & 6 & 9.750 \\
8 & 6 & \textbf{10.000} \\
\bottomrule
\end{tabular}
\caption{Hyper-parameter Tuning Results: Average High Scores}
\end{table}

\label{tab:hyperparameter_tuning}

\begin{table*}[ht]
\centering
\begin{tabular}{l l l l}
\toprule
\textbf{Model Name} & \textbf{Version} & \textbf{Parameters} & \textbf{HuggingFace Link} \\
\midrule
Hermes      & 3             & 8B   & \href{https://huggingface.co/NousResearch/Hermes-3-Llama-3.1-8B}{NousResearch/Hermes-3-Llama-3.1-8B} \\
Zephyr      & Beta          & 7B   & \href{https://huggingface.co/HuggingFaceH4/zephyr-7b-beta}{HuggingFaceH4/zephyr-7b-beta} \\
Meta-Llamaz & 3             & 8B   & \href{https://huggingface.co/NousResearch/Hermes-3-Llama-3.1-8B}{NousResearch/Hermes-3-Llama-3.1-8B} \\
Llama       & 3 Instruct    & 8B   & \href{https://huggingface.co/meta-llama/Meta-Llama-3-8B-Instruct}{meta-llama/Meta-Llama-3-8B-Instruct} \\
Nous Hermes & 2             & 7B   & \href{https://huggingface.co/NousResearch/Nous-Hermes-2-Mixtral-8x7B-DPO}{NousResearch/Nous-Hermes-2-Mixtral-8x7B-DPO} \\
Llama       & 3.1           & 70B  & \href{https://huggingface.co/meta-llama/Llama-3.1-70B-Instruct}{meta-llama/Llama-3.1-70B-Instruct} \\
Qwen        & 2.5 Instruct  & 72B  & \href{https://huggingface.co/Qwen/Qwen2.5-72B-Instruct}{Qwen/Qwen2.5-72B-Instruct} \\
\bottomrule
\end{tabular}
\caption{Model used for CS generation.}
\label{tab:models}
\end{table*}

\end{document}